\pdfoutput=1

\documentclass[11pt]{article}

\usepackage[]{EACL2023}

\usepackage{times}
\usepackage{latexsym}
\usepackage{amsmath}
\usepackage{graphicx}
\usepackage{caption}
\usepackage{booktabs}
\usepackage{pifont}
\usepackage{subfigure}
\usepackage{bbm}
\usepackage{amsfonts}
\usepackage{multirow}

\usepackage[T1]{fontenc}

\usepackage[utf8]{inputenc}

\usepackage{microtype}

\usepackage{inconsolata}

\usepackage{hyperref}
\usepackage{booktabs}
\definecolor{ao(english)}{rgb}{0.0, 0.5, 0.0}

\newcommand{\green}[1]{{\textcolor{ao(english)}{#1}}}

\title{FVQA 2.0: Introducing Adversarial Samples into Fact-based Visual Question Answering}

\author{Weizhe Lin\\
  Department of Engineering\\
  University of Cambridge \\
  United Kingdom \\
  \texttt{wl356@cam.ac.uk}\\
  \And
  Zhilin Wang\\
  Department of Linguistics\\
  University of Washington \\
  United States\\
  \texttt{zhilinw@uw.edu} \\\And
  Bill Byrne \\
  Department of Engineering\\
  University of Cambridge \\
  United Kingdom \\
  \texttt{bill.byrne@eng.cam.ac.uk} \\}

\begin{document}
\maketitle
\begin{abstract}
The widely used Fact-based Visual Question Answering (FVQA) dataset contains visually-grounded questions that require information retrieval using common sense knowledge graphs to answer.  It has been observed that the original dataset is highly imbalanced and concentrated on a small portion of its associated knowledge graph.
We introduce FVQA 2.0 which contains adversarial variants of test questions to address this imbalance.
We show that systems trained with the original FVQA train sets can be vulnerable to adversarial samples and we demonstrate an augmentation scheme to reduce this vulnerability without human annotations.

\end{abstract}

\section{Introduction}
\label{sec:FVQA:introduction}

Knowledge-based Visual Question Answering (KB-VQA) lies at the intersection of Computer Vision, Natural Language Processing, and Information Retrieval.
A KB-VQA system must access external knowledge sources to find a correct and complete answer, a task that is sometimes hard for humans.

Fact-based Visual Question Answering (FVQA)~\cite{wang2017fvqa} is a VQA task in which visually-grounded questions and answers about images are grounded by knowledge-graph (KG) triplets taken from several `common sense' knowledge bases, such as ConceptNet~\cite{speer2017conceptnet}, Webchild~\cite{tandon2017webchild}, and DBpedia~\cite{auer2007dbpedia}.
For instance, ``Question: Which thing in the image can be used for scooping food? Answer: spoon'' is associated with the KG triplet ``spoon - UsedFor - scooping food''.
These questions are challenging in that retrieving information from external KGs is necessary.

The original FVQA dataset~\cite{wang2017fvqa} has several readily observed limitations.
First, the dataset is small (5486 samples) and the annotations are limited to a single answer per question,  ignoring other correct answers. 
This limitation arises from the FVQA creation process in which annotators were first asked to select a KG triplet on which they would ask a question about an image. 
This approach prevented the annotators from labeling other valid KG triplets.
Secondly, the dataset is highly imbalanced.
Some triplets and answers are frequently used, but other KG triplets and answers are severely underrepresented in training.
For example,
there are 1,129 possible answers in total, but over 90\% of questions focus on only a half of them;
792 (70\%) answers appear less than 3 times; only 4,216 out of $\sim$220k triplets are used.

These limitations lead to a potential problem: KB-VQA systems trained on this dataset overfit on these frequently used triplets and perform poorly on variants that contain other valid triplets or other images.
Also, extensive overlap between training and test can lead to an unrealistically high question answering baseline performance.
We noted that a question with a triplet unseen in training is often answered with `person', since it is the most frequent answer in the original data distribution.

To overcome these limitations, we introduce an enlarged test set that contains two types of adversarial samples (as shown in Fig.~\ref{fig:FVQA:workflow}):
\begin{figure*}[!htp]
    \centering
    \includegraphics[width=\textwidth]{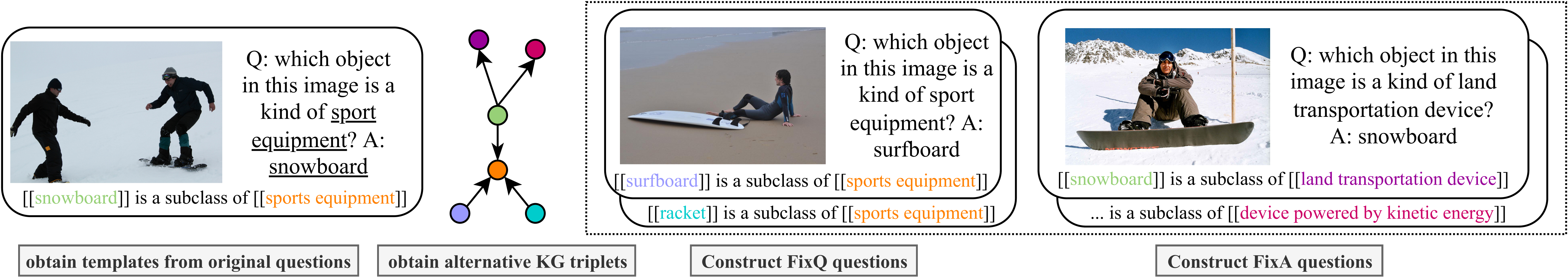}
    \vspace{-0.7cm}
    \caption{The workflow of constructing adversarial samples (FixQ and FixA questions) from the original test set questions.}
    \label{fig:FVQA:workflow}
    \vspace{-0.5cm}
    
\end{figure*}
(1) \textit{FixQ}: the question remains the same, but is associated with a different image and a different correct answer.
This ensures that a system is less able to achieve high performance if it is biased by language patterns in questions;
(2) \textit{FixA}: the answer remains the same, but the question is asked in a different way.
This favours systems that do more than make straightforward associations between questions and answers based on the training data.
In contrast to the original test set,
this new set further challenges KB-VQA system to retrieve knowledge from KBs and answer questions without being biased towards frequent answers in the original dataset.
We show that models trained on the original FVQA training sets are significantly less robust on these adversarial test samples.

Given that it is hard to guarantee a good triplet coverage during annotation,
we explore an augmentation scheme to address this problem without costly human annotation of large-scale adversarial training samples.
Our scheme generates slightly noisy adversarial samples that improve the coverage of valid KG triplets to enhance model training.

Our contributions are:

(1) We introduce FVQA 2.0, which adds an adversarial test set that challenges KB-VQA system robustness to adversarial variants of questions.

(2) We demonstrate the performance gap between the original test set and the adversarial test set, showing that considering adversarial samples is important for better realistic KB-VQA performance.



(3) To further demonstrate the importance of adversarial samples, we leverage a semi-automated augmentation scheme to improve system robustness on the adversarial test through the creation of large-scale noisy adversarial examples.

\section{Related Work}

KB-VQA questions can focus on facts and concepts, as in FVQA~\cite{wang2017fvqa} and OK-VQA~\cite{marino2019ok}.
Such questions challenge the information retrieval ability of systems.
KB-VQA questions can also require commonsense reasoning, as in parts of OK-VQA and A-OKVQA~\cite{schwenk2022okvqa}.
In particular, S3VQA~\cite{jain2021select} is an augmented version of OKVQA, improving both the quantity and quality of some question types.
A-OKVQA has shifted its core task to ``reasoning questions''. Only 18\% of questions in A-OKVQA require answers from an external knowledge base.

VQA 2.0~\cite{goyal2017making} collects `complementary images' such that each question is associated with a pair of images that result in different answers.
\citet{jain2021select} derive new S3VQA questions from manually defined question templates.
They annotated spans of objects that could be replaced, and then substituted them with a complicated substitute-and-search system.
In contrast to their labour-intensive annotation work, our adversarial samples are collected through a semi-automatic approach that fully leverages the structural information in KGs to significantly reduce the human work required.

More broadly, in Knowledge-Graph Question Answering (KG-QA), work has exploited KG to generate synthetic data in unseen domains~\cite{Trond2020neural, trivedi2017lc, linjordet2020sanitizing}.
Our work extends visually-grounded questions with valid common sense KG triplets.

\section{Method}
\label{sec:FVQA:method}

\textbf{Extracting Question Templates.}
We extract question templates that can be used to reconstruct new questions using other valid KG triplets.
We apply a rule-based system to replace KG entities that appear in the questions.
For example, `what is used for storing liquid in this image?' is transformed to `what is used for \texttt{<t>} in this image?' given that the associated KG triplet is ``bottle (\texttt{<h>}) - /r/UsedFor (\texttt{<r>}) - store liquid (\texttt{<t>})''.

For each template, we construct new question-answer pairs by exploring the node structure of the KG.
For example, ``bottle - /r/UsedFor - hold water'' is also a valid triplet from ConceptNet, whose head and relation are the same as the original triplet.
A new question ``Q: what is used for holding water in this image? A: bottle'' can now be constructed.

\textbf{Template Filtering.} We focus on questions about object concepts that are transferable to other images, ignoring a small portion ($<$10\%) of FVQA questions to which the answers are based on particular scenes (e.g. `what can you often find in the place shown in this picture?').

Human annotators are employed to filter out non-transferable templates, such as questions that contain specific object positioning (``what is the object in the lower right of this image used for?'').
This process takes around 1 hour with two annotators to obtain 440 valid templates after removing highly similar templates.

\textbf{Matching Suitable Images.}
We use 619 of FVQA images\footnote{FVQA images are from Microsoft COCO~\cite{cocodataset2014} and ILSVRC~\cite{ILSVRC15}.} that are also present in the Visual Genome dataset~\cite{visualgenomedataset2017}.
Using the object annotations of the VG dataset to determine if an image contains the object being asked, we employ a rule-based system to assign a suitable image to each generated adversarial sample, within which process all images are assigned to approximately the same number adversarial samples by a simple approach described in Appendix~\ref{sec:appendix:balance_algorithm}.
We limit the number of FixQ and FixA questions generated by each template to 5, which guarantees a reasonable dataset size.
3,805 questions are generated.

\textbf{Manual Verification.}
We conduct manual verification to rule out samples that are incorrectly generated.
432 counter-intuitive KG triplets are removed in this step.
Finally, we obtain 2,820 adversarial samples, offering 1,671 new valid triplets from the KG. Around 75\% samples are verified as correct, showing that the rule-based generation works well.
The remainder are discarded.

The official FVQA evaluation performs 5-fold validation: each split preserves around half its samples for testing.
As a naming convention, under each split, the templates extracted from the original training samples are called `train templates' while the rest are `test templates'.
Since the train templates may contain language patterns that have been learned in training, we ensure that only questions derived from test templates are used in the adversarial testing.
As a result, we have 1,376 adversarial test samples per split on average, with 1,129 FixQ and 246 FixA questions.

\textbf{Augmentation with Adversarial Data.}
We explore an augmentation scheme to augment the training data with slightly noisy but auto-generated adversarial samples, which avoids heavy annotation work.
In each split, \textbf{only the train templates} (defined in the above paragraph) are used to generate adversarial samples for training such that no information of test samples is leaked to training.
This avoids biasing the training to the test sets, which would make the test sets less indicative of true system performance.
We obtain an augmentation set with 2,262 questions per split on average semi-automatically, which would otherwise cost hundreds of hours to build from scratch.
The origins of these adversarial samples are referred to as `\textit{Originating Questions}'.
There are 435 such questions per split.
In training, these questions are randomly replaced by their adversarial variants.

\section{FVQA 2.0 Statistics}
\label{sec:appendix:dataset_statistics}
\begin{table}[!ht]
\centering
\small
\begin{tabular}{lrr}
\toprule
Set Name             & \#Samples & std \\
\midrule
Standard Train Set   & 2,927    & 69   \\
Standard Test Set    & 2,899    & 69   \\
\midrule
Originating Questions Set & 435     & 52   \\
Adversarial Test Set & 1,376    & 193  \\
~~~~- FixA Questions       & 1,129    & 157  \\
~~~~- FixQ Questions       & 246     & 38   \\
\midrule
Augmentation data    & 2,262    & 267  \\
\bottomrule
\end{tabular}
\caption{Dataset Statistics. \#Samples: average number of samples across 5 folds; std: the standard deviation over 5 folds.}
\label{tab:FVQA:dataset_statistics}
\end{table}

The numbers of samples in each set are provided in Table~\ref{tab:FVQA:dataset_statistics}.
The official FVQA dataset creates 5 folds by splitting the images being used.
Half of these images are used in training while the other half are reserved for testing.
In all our new sets, under each split, questions for training are not leaked to testing.
The `Originating Question Set' is a subset of Standard Test Set by its definition (Sec.~\ref{sec:FVQA:method}).
The Adversarial Test Set is formed by FixA questions and FixQ questions;
it is created by automatically generating adversarial question variants from the questions in the `Originating Question Set'.
It covers relationships such as /r/RelatedTo, /r/IsA, /r/PartOf, /r/HasA, /r/UsedFor, /r/CapableOf, /r/AtLocation, /r/Desires, /r/MadeOf.
The augmentation data consists of adversarial variants that are derived from the questions in the Standard Train Set.

\section{Experiments}
\textbf{Baseline Systems}
We use several FVQA systems for comparison\footnote{Since many recent FVQA systems are not open-sourced, we additionally include KB-VQA systems from OKVQA.}:
FVQA~\cite{wang2017fvqa}, the baseline system provided in the official FVQA dataset paper;
GCN~\cite{fvqa_gcn}, a model that leverages graph convolutional networks (GCNs) to aggregate features from visual/language/fact modalities;
Mucko~\cite{zhu2020mucko}, the current state-of-the-art system that uses GCNs to combine visual, fact, and semantic graphs. 

We test our augmentation scheme on several systems that have code available:
\textbf{RAVQA-NoDPR} and \textbf{RAVQA-DPR}~\cite{lin-etal-2022-retrieval}, T5~\cite{t5paper}-based models that transform images into texts (e.g. objects, attributes, and image captions) and the DPR version additionally uses Dense Passage Retrieval~\cite{karpukhin-etal-2020-dense} to retrieve documents from knowledge bases\footnote{In our experiments, the knowledge base consists of surface texts of triplets (e.g. ``[car] has [4 wheels]'').};
\textbf{TRiG}~\cite{gao2022transform}, a model that is similar to RAVQA-DPR but different in embedding fusion;
\textbf{ZS-F-VQA}~\cite{chen2021zero}, an FVQA system that obtains the final prediction by fusing the individual predictions in answer/fact/relation graphs.

\textbf{Metrics.} We report accuracy and standard deviation over 5 splits (Sec.~\ref{sec:appendix:dataset_statistics}).
In calculating accuracy for open-ended generation systems (RAVQA/TRiG), 
a question is considered successfully answered if the generated answer string is an exact match to the ground-truth answer node, which is the closest KG node to the ground-truth answer string (shortest in Levenshtein distance computed from node names).

\textbf{Performance and Discussion.}
\begin{table*}[!htp]
\small
\centering
\begin{tabular}{lccccccc}
\toprule
Test on:    & \multicolumn{2}{c}{Standard Test Set} & \multicolumn{2}{c}{Originating Question Set} & \multicolumn{2}{c}{Adversarial Test Set}  \\ \cmidrule(r){2-3} \cmidrule(r){4-5} \cmidrule(r){6-7}
Trained on: & Original          & Augmented         & Original            & Augmented         & Original*         & Augmented (\green{improv. over *})        \\
\cmidrule(r){0-0}
\cmidrule(r){2-3} \cmidrule(r){4-5} \cmidrule(r){6-7}
ZS-F-VQA    & 48.16 {\tiny $\pm$1.03}      & 48.57 {\tiny $\pm$1.00}   & 63.67 {\tiny $\pm$0.88}     & 64.63 {\tiny $\pm$0.81}   & 49.97 {\tiny $\pm$2.37}        & 74.06 {\tiny $\pm$1.92} ~\green{+24.09}            \\
TRiG  &  64.94 {\tiny $\pm$0.93}& 	65.73 {\tiny $\pm$0.33}& 	81.67 {\tiny $\pm$1.12}& 	83.48 {\tiny $\pm$1.89}& 	68.86 {\tiny $\pm$3.26}& 	79.79 {\tiny $\pm$1.34}~\green{+10.93}
\\
RAVQA-NoDPR & 66.19 {\tiny $\pm$1.15}      & 66.70 {\tiny $\pm$1.00}   & 84.59 {\tiny $\pm$1.24}     & 85.75 {\tiny $\pm$0.90}   & 71.48 {\tiny $\pm$2.08}        & 82.38 {\tiny $\pm$1.65}~\green{+10.90}            \\
RAVQA-DPR & 69.56 {\tiny $\pm$0.78}      & 69.90 {\tiny $\pm$0.56}      & 87.52 {\tiny $\pm$1.68}     & 88.33 {\tiny $\pm$1.40} & 76.91 {\tiny $\pm$1.93}        & 85.05 {\tiny $\pm$1.15}~~~\green{+8.14}           \\ \bottomrule
\end{tabular}
\vspace{-0.2cm}
\caption{Model performance on the standard test set, originating questions (from which the adversarial questions are derived), and adversarial test set. Results are reported as the average of 5 folds with standard deviations.}
\label{tab:FVQA:performance}
\end{table*}
Table~\ref{tab:FVQA:performance} shows that the systems used for evaluating the new adversarial set are sufficiently strong (e.g. 69.56\% accuracy by RAVQA-DPR) in comparison with the three models that do not have code available, which achieve 58.76\% (FVQA), 69.35\% (GCN), and 73.06\% (Mucko, current state-of-the-art) respectively.
RAVQA-NoDPR achieves 84.59\% accuracy on the originating questions but obtains only 71.48\% accuracy on the adversarial samples derived from them.
Such performance gaps are readily observed on all systems.
Systems trained on the original training sets fail to perform equally well on the two sets, showing that the original FVQA training data does not contain adversarial variants and the resulting systems are vulnerable to them.

By incorporating adversarial variants in training, all systems achieve much better performance on the challenging adversarial set, e.g. RAVQA-NoDPR is improved from 71.48\% to 82.38\% (+10.9\%).
The performance on the standard and adversarial test sets now match well, with the gap reduced from more than 10\% to $\sim$3\%, showing that the augmentation scheme significantly improves systems' reliability and robustness.
The relative improvement is slightly less (+8.1\%) for RAVQA-DPR, which is expected given that it is a retrieval-based system designed to answer both seen and unseen questions with its strong retrieval ability.
ZS-F-VQA benefits greatly from augmentation: its adversarial performance is improved by 24.09\%.
This is because its model size is much smaller and it can easily be biased by language patterns, images, and frequent answers seen in training.

In summary, systems trained on the original training sets are vulnerable to adversarial variants of the test questions.
We show that through generating adversarial samples for data augmentation, systems become much more robust to these variants.

\textbf{Analysis of Model Vulnerability.}
\begin{figure}
    \centering
    \includegraphics[width=0.8\linewidth]{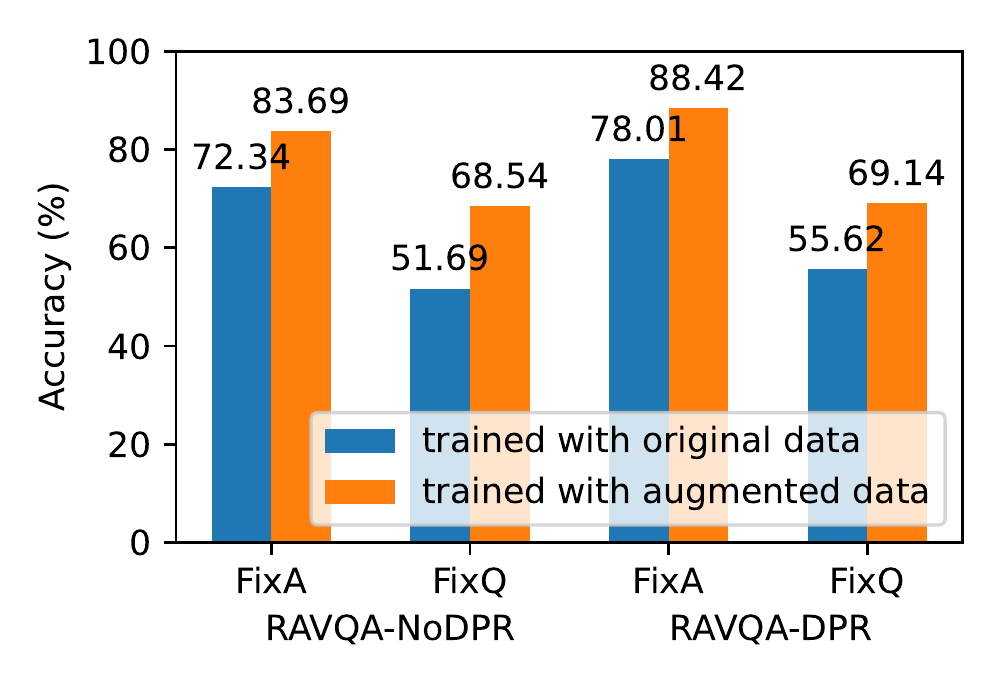}
    \vspace{-0.5cm}
    \caption{Performance on FixQ and FixA questions.}
    \label{fig:FVQA:question_type_analysis}
    \vspace{-0.2cm}
\end{figure}
As shown in Fig.~\ref{fig:FVQA:question_type_analysis}, RAVQA systems trained with original training sets perform better on FixA questions ($\sim$88\%) than on FixQ questions ($\sim$69\%).
This suggests that systems perform worse when asked the same questions on different images.
This is potentially because the language patterns seen in  training bias the models to frequent choices, lowering the FixQ generalizability.
In contrast, systems are less distracted by different ways of asking for the same answer, potentially due to the strong language modelling capability of T5 used by them.
The augmentation scheme improves systems on both types of questions significantly (by $\sim$10\% on each), showing the value of adversarial samples in training.

\begin{figure}
    \centering
    \includegraphics[width=0.98\linewidth]{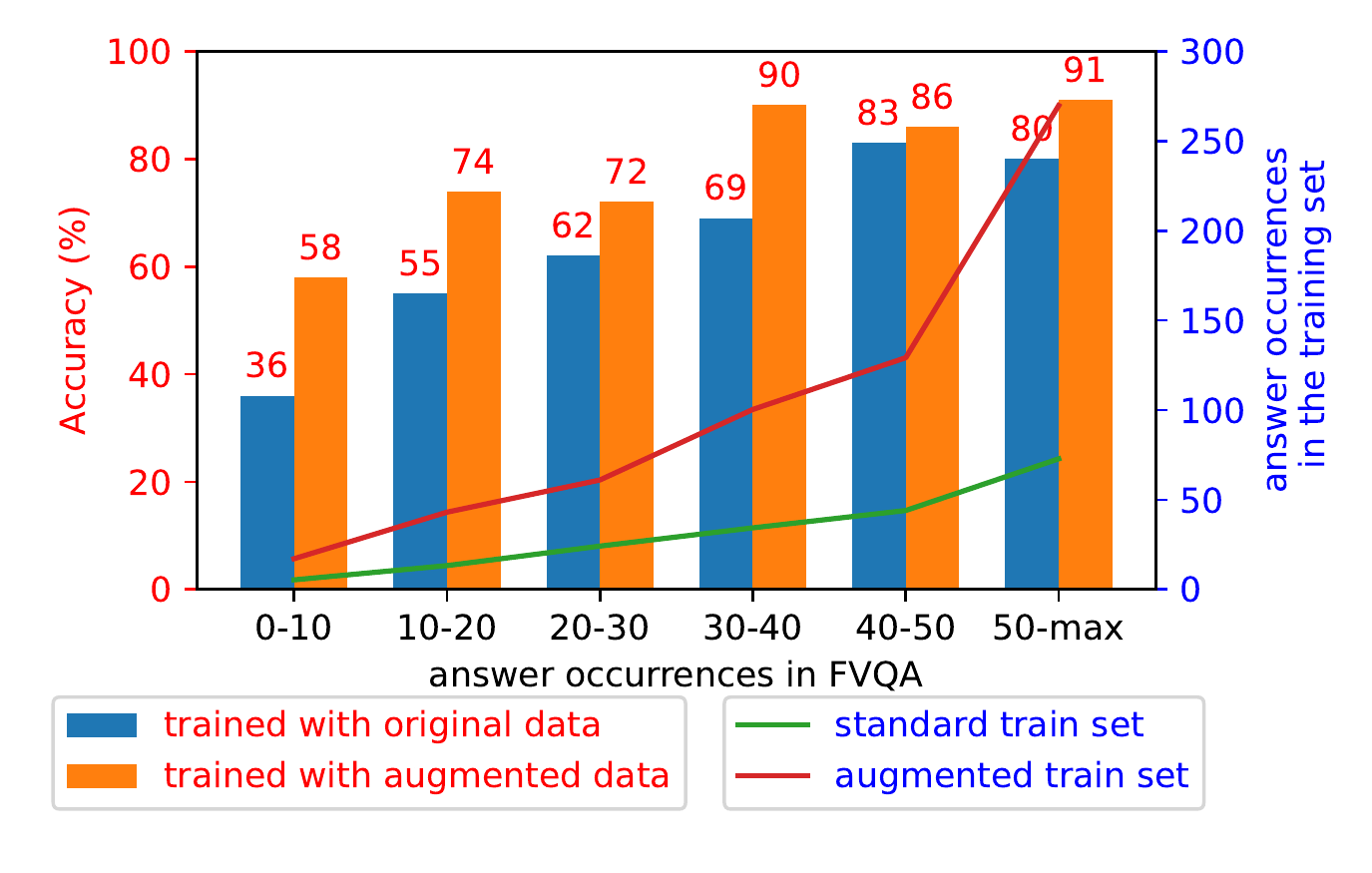}
    \vspace{-0.2cm}
    \caption{RAVQA-DPR accuracy on adversarial questions and answer occurrences in the standard/augmented training sets.
    They are grouped by the number of answer occurrences in the original FVQA dataset. For example, a question is counted towards the `0-10' group if its answer appears less than 10 times in the original dataset.}
    \label{fig:FVQA:data_distribution}
    \vspace{-0.3cm}
\end{figure}

Fig.~\ref{fig:FVQA:data_distribution} plots the RAVQA-DPR performance on the adversarial test set questions that are grouped by their answer occurrences in the original FVQA dataset.
The answer distribution of the original dataset affects adversarial performance greatly: systems perform much worse on questions whose answers appear less frequently in FVQA.
In contrast, the performance deterioration that arises from answer rarity is mitigated significantly after augmentation.
The augmentation scheme (red v.s. green curve in Fig.~\ref{fig:FVQA:data_distribution}) compensates for the imbalanced answer distribution by providing more question variants so that systems are trained on both popular and rare answers.

\section{Conclusion}

We show that the FVQA test sets are not sufficiently indicative of true system performance through providing a new human-verified adversarial test set that contains adversarial variants of the original test set questions.
We show the value of adversarial samples in KB-VQA datasets by showing an augmentation scheme that leverages structural information in KGs to create augmentation questions for training, which improves models' robustness to adversarial variants.


We release the dataset and the codes in Github (\href{https://github.com/LinWeizheDragon/Retrieval-Augmented-Visual-Question-Answering}{https://github.com/LinWeizheDragon/Retrieval-Augmented-Visual-Question-Answering}).

\section{Limitations}

The adversarial test set was firstly generated from the original FVQA dataset by a rule-based system and then filtered by human annotators.
As a result, the new set is limited with respect to the question types, language patterns, and knowledge triplets used in FVQA.
One potential solution to overcome this limitation is to invest more human effort to generate adversarial questions from scratch, which is, however, much more expensive than the semi-automatic approach presented here.

The proposed augmentation approach also relies on the relationships encoded in the knowledge base (e.g. ConceptNet~\cite{speer2017conceptnet}).
These will influence the quality and diversity of the augmented data, with the expectation that improvements in KG scope and quality will improve data augmentation.

The number of adversarial examples introduced in this work is sufficiently large for investigating the performance discrepancies (on the original and adversarial test sets) and demonstrating the necessity of KB-VQA adversarial samples.
However, it is considered beneficial to introduce adversarial samples on a larger scale by considering them in the design of future KB-VQA datasets.

\section{Ethics Statement}
Our dataset was created semi-automatically from the FVQA dataset and ConceptNet, a crowd sourced common sense knowledge graph.
Though we have included human annotators in the loop to remove sexual, offensive, and other inappropriate data samples that were automatically generated (we removed $\sim$200 inappropriate knowledge graph triplets during annotation), we recognize that the dataset may still contain a small number of inappropriate samples. Any developers who replicate the semi-automatic methodology described in the paper to extend the datasets should include a similar review step in the manual work flow.
We also recognize that the systems trained on this dataset may convey such inappropriate information to users in real-life applications.
Therefore, extra care must be taken when using this dataset in applications that interact directly with real users.

\section{Acknowledgement}
W. Lin was supported by a Research Studentship funded by Toyota Motor Europe (RG92562(24020)). We thank our colleagues, Daniel Olmeda Reino (Toyota Motor Europe) and Jonas Ambeck (Toyota Motor Europe), who provided insight and expertise in this project.

We thank Alexandru Coca (University of Cambridge) for comments that greatly improved the manuscript.
We would also like to thank all the reviewers for their knowledgeable reviews.

\bibliography{anthology,paper_ref}
\bibliographystyle{acl_natbib}

\appendix

\section{Training Details}
\textbf{ZS-F-VQA}: The experiments were performed on 1 $\times$ Nvidia RTX 3090. We used the code from the official repository\footnote{\href{https://github.com/China-UK-ZSL/ZS-F-VQA}{https://github.com/China-UK-ZSL/ZS-F-VQA}}.
The original paper dropped questions that have rare answers.
For fair comparison with other models, we added these rare answers back and performed training and testing.
We chose to report the performance of the system which uses `SAN' as the base model (details are in the paper and the repository), since this setting has achieved the best performance.
The hyperparameters for training are kept the same as the original paper. In testing, we selected $k_e=10; k_r=1; score=10$ by grid search (search range: $0\leq k_e\leq 20; 0\leq k_r\leq 20; 0\leq score\leq 20$).

\textbf{RA-VQA-NoDPR/RAVQA-DPR/TRiG}:
All experiments were performed on 1 $\times$ Nvidia A-100 GPU.
We chose Adam~\cite{KingmaB14adam} as the optimizer.
When the model has a DPR component, we trained the DPR component for 4 epochs with a constant learning rate $10^{-5}$.
In training the answer generator, the learning rate linearly decays from $6 \times 10^{-5}$ to $0$ after $10$ epochs, as suggested in the original paper.
For each split, the checkpoints at global step 2k (around 3.5 epochs) were used in testing.
We retrieve 5 best documents when predicting the answer ($K_{\text{train}}=5$), since this number was reported to best balance the computation and performance \cite{lin-etal-2022-retrieval}.

We obtained the pre-trained model parameters (T5-large and BERT-base) from Huggingface~\cite{wolf-etal-2020-transformers}.
These systems are implemented with Huggingface Python libraries (under Apache License 2.0).
The FAISS~\cite{johnson2019billion} system is under MIT License.

\section{Balancing Images in Adversarial Variants}
\label{sec:appendix:balance_algorithm}
In assigning suitable images to question templates, it is necessary to ensure the diversity of images being used.
We achieve this by controlling the number of assignments per image with a simple approach so that the numbers are approximately the same for all images.

In the process, for each new question-answer pair that needs an image, we rank all the images that contain the object being asked in the the question by their current total number of assignment.
We select the image that satisfies the conditions as well as having the fewest number of assignment as the associated image of the new sample.
We found that by applying this simple yet effective strategy, the assigned images present a good diversity.

\section{Annotation Details}
Two annotators (volunteers in the research group) worked independently to rule out incorrectly generated examples.
An example was accepted only if the two annotators achieved consensus.
The annotators attempted to fix grammar errors that caused severe misunderstanding, while mild errors were kept (for example, `is used for carry people' does not prevent models/people from understanding the question, and thus the annotators are not required to fix them).

In particular, questions that might contain information of individuals / private information were dropped, though it is a very rare case.

\textbf{Questions with multiple answers:} when a question can be answered with multiple instances in an image, all possible answers are included. During annotation, incorrect answers were dropped from the list. In evaluation, answering any correct answer is considered successful. There are around 11\% multiple-answer questions at the end.

\section{Additional Results}

We include some additional baseline performance in Table~\ref{tab:FVQA:appendix:additional-performance}.
It can be easily seen that the performance on originating questions (the original FVQA questions that are used to derive the adversarial samples) is very high even when images are excluded. This further supports our argument that the original dataset is heavily biased to frequent answers. The performance on the adversarial set is lower, showing that this new test set is more challenging and less biased toward language patterns.

\begin{table*}[!htp]
\small
\centering
\begin{tabular}{lccc}
\toprule
Models    & \multicolumn{1}{c}{Standard Test Set} & \multicolumn{1}{c}{Originating Question Set} & \multicolumn{1}{c}{Adversarial Test Set}  \\
\midrule
RAVQA-DPR & 69.56 {\tiny $\pm$0.78}      &  87.52 {\tiny $\pm$1.68}    & 76.91 {\tiny $\pm$1.93}                   \\
~~~~~\textit{(without triplets)} & 66.19 {\tiny $\pm$1.15}        & 84.59 {\tiny $\pm$1.24}       & 71.48 {\tiny $\pm$2.08}        \\
~~~~~\textit{(without images)} & 43.83 {\tiny $\pm$0.68}      & 57.53 {\tiny $\pm$2.93}   & 50.02 {\tiny $\pm$1.00}            \\
~~~~~\textit{(without triplets and images)} & 40.29 {\tiny $\pm$1.60}      & 51.41 {\tiny $\pm$3.25}   & 42.55 {\tiny $\pm$0.90}            \\
\bottomrule
\end{tabular}
\vspace{-0.2cm}
\caption{The performance of some additional baseline systems on the standard test set, originating questions (from which the adversarial questions are derived), and adversarial test set. Results are reported as the average of 5 folds with standard deviations.}
\label{tab:FVQA:appendix:additional-performance}
\end{table*}

\section{More Examples of FVQA 2.0}
\begin{figure*}[!h]
    \centering
    \includegraphics[width=\textwidth]{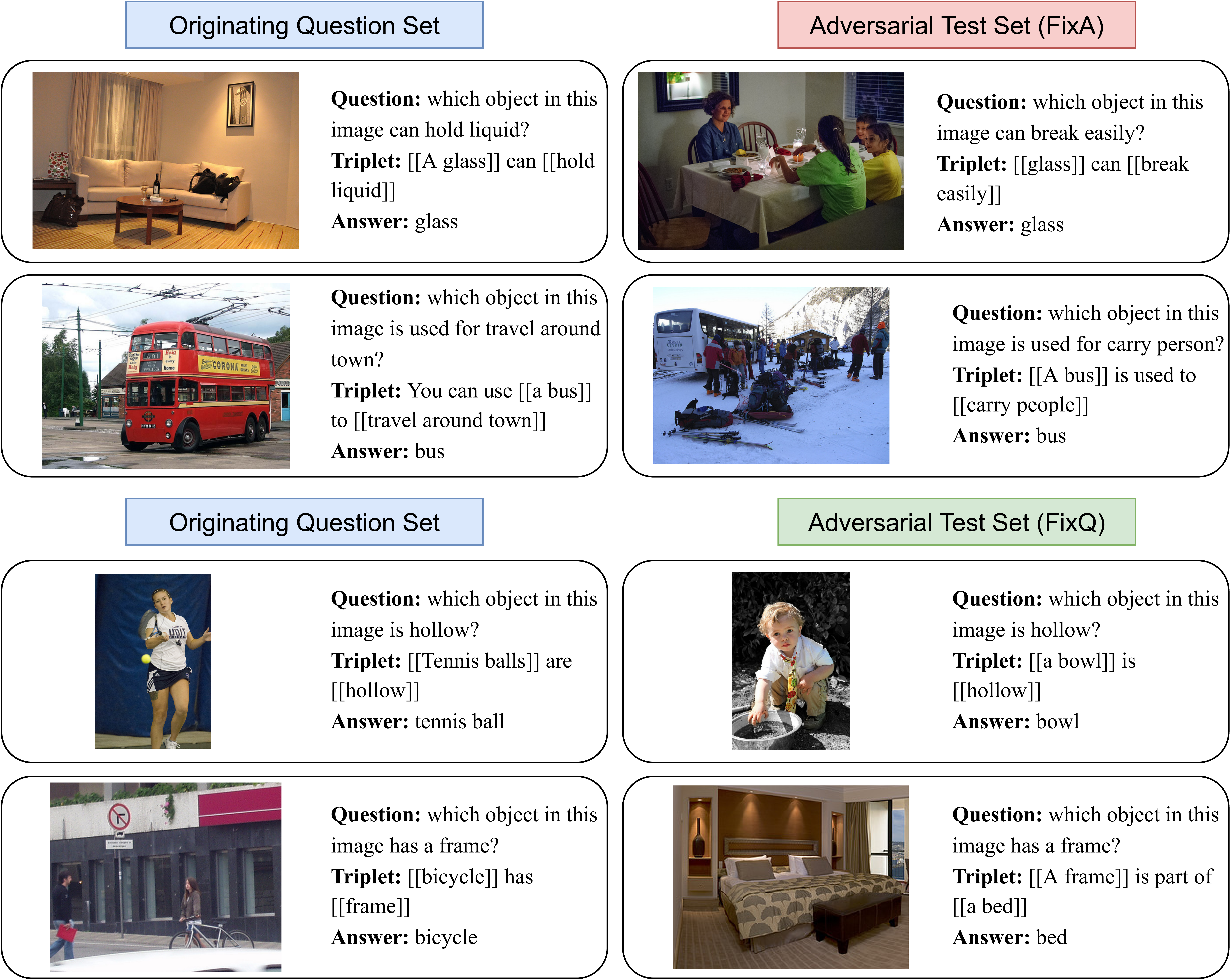}
    \caption{More examples taken from the FVQA 2.0 adversarial test set. The questions in the left column are from the official FVQA test set. They are used to derive the adversarial questions in the right column. FixA: the answer remains the same while the way of asking for the answer is different; FixQ: the question remains the same, but the answer changes in a different image. More details are presented in Sec.~\ref{sec:FVQA:introduction}.}
    \label{fig:FVQA:examples}
\end{figure*}

We demonstrate some more examples from the new Adversarial Test Set in Fig.~\ref{fig:FVQA:examples}.

\end{document}